\documentclass{article}

\PassOptionsToPackage{numbers,sort&compress}{natbib}
\usepackage[preprint]{neurips_2025}
\makeatletter\renewcommand{\@noticestring}{}\makeatother

\usepackage[utf8]{inputenc}
\usepackage[T1]{fontenc}

\usepackage{hyperref}
\usepackage{url}
\newcommand{\doi}[1]{\href{https://doi.org/#1}{doi: \nolinkurl{#1}}}
\usepackage{booktabs}
\usepackage{amsmath}
\usepackage{amsfonts}
\usepackage{nicefrac}
\usepackage{microtype}
\usepackage{xcolor}
\usepackage{graphicx}
\usepackage{array}
\usepackage{tabularx}
\usepackage[section]{placeins}

\hypersetup{
  colorlinks=true, linkcolor=blue!60!black, citecolor=blue!60!black, urlcolor=blue!60!black,
  pdftitle={SECURE-AI: Multi-Signal Uncertainty Quantification for Trustworthy Acoustic Cough Detection},
  pdfauthor={Sejal Solkhan}
}

\title{SECURE-AI: Multi-Signal Uncertainty Quantification\\for Trustworthy Acoustic Cough Detection}

\author{%
  Sejal Solkhan\textsuperscript{1}, Feiya Xiang\textsuperscript{2}, Mayur Sanap\textsuperscript{2}, Geet Khatri\textsuperscript{2}, and Edgar Lobaton\textsuperscript{2}
}

\begin{document}

\maketitle
\makeatletter
{\renewcommand{\@makefntext}[1]{\noindent #1}\footnotetext{\textsuperscript{1}Green Level High School, NC, USA. E-mail: \texttt{sejal.solkhan.research@gmail.com}.\\
\textsuperscript{2}Department of Electrical and Computer Engineering, North Carolina State University, Raleigh, NC 27695, USA. Corresponding author e-mail: \texttt{sejal.solkhan.research@gmail.com}.}}
\makeatother

\begin{abstract}
Acoustic models for cough detection achieve high accuracy on curated benchmarks but fail silently in deployment, where speech, environmental noise, and audio unlike the training data induce confident misclassifications. These are especially consequential in health monitoring, where a confidently wrong prediction is worse than no answer at all. We present SECURE-AI (Self-aware Estimation of Cough Uncertainty for Reliable Evaluation of AI), a multi-signal uncertainty quantification framework that wraps a frozen, pre-trained MobileNetV3 cough classifier and estimates when its predictions should not be trusted, combining Monte Carlo Dropout, Temperature Scaling, and Mahalanobis distance in the embedding space. On a held-out split of 12,161 Coswara recordings, the probability-space methods are individually insufficient: Monte Carlo Dropout assigns lower uncertainty to incorrect than to correct predictions, and temperature scaling improves calibration without improving discrimination. The strongest signal is embedding-space distance: true positives and false positives separate in Mahalanobis distance (true-positive mean 44.7 versus false-positive mean 63.4; Welch $t = -23.25$, $p < 0.001$; Cohen's $d = 0.94$), with an area under the receiver operating characteristic curve of 0.86. Among predictions classified as non-coughs, lower Mahalanobis distance also separates false negatives from true negatives, with an area under the curve of 0.82. Thus, the same embedding signal can help identify missed coughs as well as false alarms. At a descriptive operating point selected from labeled test data, retaining 90 percent of baseline true-positive detections removes about two-thirds of false positives and raises precision from 59.9 percent to roughly 80 percent. We also find that a rejection threshold tuned on one data split does not transfer to another (validation F1 of 82 percent falls to 34 percent on test), so we report threshold-independent area under the curve as the main performance measure. An error taxonomy shows that failures are systematic, with most false positives related to speech, and that different error classes respond to different uncertainty signals. The pipeline runs on Raspberry Pi hardware costing 30 to 90 US dollars. These results indicate that probabilistic calibration alone is inadequate for reliable cough detection and that embedding-space analysis can improve the detection of both false alarms and missed coughs.
\end{abstract}

\section{Introduction}

Cough is the symptom for which patients most frequently seek medical attention, and a principal transmission mechanism for respiratory infectious disease including tuberculosis, influenza, and COVID-19. In chronic conditions such as asthma, COPD, and interstitial lung disease, cough frequency functions as a clinically meaningful biomarker of disease progression and treatment response; increases in cough frequency can precede acute exacerbations by several days, creating an early-intervention window that occasional clinic visits routinely miss.

Yet cough assessment in standard practice remains subjective. Clinicians rely on patient self-report, which is subject to recall bias, or on manual counting from 24-hour audio recordings that can be accurate but labor-intensive, costly, and not practical at scale. Automated acoustic cough detection promises to close this gap, and modern deep classifiers report accuracies above 95 percent on curated benchmarks.

In practice, these numbers do not hold in real-life scenarios. A wearable microphone in a home or clinic captures speech, laughter, door slams, television audio, and fabric rustle, sound classes largely absent from training corpora. A standard softmax classifier assigns every input to one of its known classes, and does so with high confidence, regardless of whether the input resembles its training data. The resulting failure is not a gradual loss of accuracy but a silent one: the model is confidently incorrect, and nothing in its output indicates that the prediction is unreliable.

For cough detection, the two error types are not equally costly. A missed cough is a lost data point, but a confidently reported false cough distorts the symptom trend a clinician is reading, reduces user trust, and at scale undermines the value of continuous monitoring. What matters most for real use is therefore reliability, not overall accuracy.

This paper asks whether uncertainty quantification can supply the missing signal. Concretely, we ask three questions: (i) Can uncertainty quantification improve the reliability of cough detection in real-world audio? (ii) Can it identify the specific predictions that require human verification? (iii) Can a structured error taxonomy expose actionable failure patterns? Our central hypothesis is that a multi-signal framework will catch failures that no individual method detects alone, because different failure modes are legible to different signals, and a system relying on any one of them inherits that method's blind spot.

This work makes the following contributions. The pre-trained MobileNetV3 cough classifier is provided by the Active Robotic Sensing (ARoS) Laboratory at North Carolina State University and is used unmodified, in inference mode only; it is the experimental control, not something this paper contributes. On top of it, we contribute: a multi-signal uncertainty quantification framework composing epistemic uncertainty, aleatoric uncertainty, calibration confidence, and embedding-space distance, with step-by-step tests showing what each signal adds; the finding that confident false positives (high-confidence, low-epistemic-uncertainty errors) are invisible to Monte Carlo Dropout and temperature scaling but are separable in embedding space; a structured error taxonomy linking each error class to the uncertainty signal that detects it; statistical validation with effect sizes, a threshold-independent separability measure (area under the ROC curve), and the finding that distance-based rejection thresholds do not transfer across data splits; and two wearable prototypes demonstrating uncertainty-aware inference on commodity edge hardware.

\section{Related Work}

\paragraph{Uncertainty in deep networks.}
Gal and Ghahramani~\cite{gal2016} showed that dropout at inference time approximates Bayesian inference in deep Gaussian processes and provides stochastic predictions for estimating model uncertainty. In this study, those predictions are used to calculate both epistemic and aleatoric uncertainty, as defined in Section~\ref{sec:signals}. Monte Carlo Dropout is attractive when the classifier was trained with dropout because it can be applied without retraining or changing the learned weights. In our results, however, model disagreement did not reliably identify input novelty: a network can be stably and incorrectly confident about an input unlike the training data.

\paragraph{Calibration.}
Guo et al.~\cite{guo2017} documented that modern neural networks are systematically overconfident and showed that a single scalar temperature applied to the logits before the softmax substantially improves calibration. Temperature scaling is post-hoc, cheap, and preserves the argmax. This is precisely why it cannot help here in the way one might hope. It rescales confidence; it does not change which inputs the model finds familiar. As we show in Section~\ref{sec:temp}, it also interacts badly with the fixed decision thresholds that threshold-based detection tasks require.

\paragraph{Out-of-distribution detection.}
Hendrycks and Gimpel~\cite{hendrycks2017} established the maximum-softmax-probability baseline for out-of-distribution detection. Liang et al.~\cite{liang2018} improved on it with ODIN, using temperature scaling and an input perturbation obtained from a gradient calculation, which adds computation at inference. Lee et al.~\cite{lee2018} proposed class-conditional Gaussian models of deep feature space and scored inputs using Mahalanobis distance to the nearest class mean. ODIN and the Mahalanobis method were developed mainly on image-classification benchmarks. The gap addressed here is not out-of-distribution detection in acoustic health monitoring generally, but an audio-only comparison of probability-space and embedding-space signals on the same error types.

\paragraph{Selective prediction.}
Geifman and El-Yaniv~\cite{geifman2017} applied selective classification to deep neural networks and studied the risk-coverage tradeoff for classifiers with a reject option. Their method can use different confidence-ranking scores rather than prescribing a single score. Therefore, choosing the score that drives rejection remains an important design decision and, as we show, affects real-world performance.

\paragraph{Cough detection and wearables.}
Automated cough detection has been pursued with multimodal sensing~\cite{albini2024,dixon2025}, radar~\cite{lu2025}, and audio classifiers trained on large sound-event corpora~\cite{gemmeke2017,fonseca2022,koutini2022}. Most relevant here, Chen et al.~\cite{chen2026} address out-of-distribution detection for wearable cough monitoring using multimodal input. Our work is complementary and differs in focus: we use a single audio channel and a frozen classifier, compare probability-space and embedding-space uncertainty signals on the same errors, and examine which failure types each signal detects. Combining the signals provides broader error coverage than using any one signal alone.

\paragraph{Positioning.}
Prior methods focus on different aspects of model reliability, while real-world audio can present several failure modes at once. The contribution of this paper is not a new uncertainty estimator but to show that different types of errors are detected by different signals and to identify which signal detects each error.

\section{Materials and Methods}

\subsection{Datasets}

Two public datasets were used without relabeling or data augmentation. Coswara~\cite{sharma2020} (Indian Institute of Science, Bangalore) comprises 24,322 crowd-sourced respiratory audio samples, of which 5,408 (22.2 percent) are coughs. Recordings were collected in uncontrolled conditions across heterogeneous devices and environments, and the dataset contains substantial speech, breathing, and environmental audio. It is our primary evaluation set precisely because it is unclean. ESC-50~\cite{piczak2015} comprises 2,000 environmental audio clips manually extracted from public field recordings and organized into balanced sound classes, of which 40 (2 percent) are coughs. It is curated and class-balanced rather than a controlled laboratory recording, and serves as a clean, well-separated contrast to the noisier Coswara data. Together the two corpora bracket the deployment problem: ESC-50 measures behavior on clean, well-separated sound classes; Coswara measures behavior on the messy, imbalanced, noise-heavy audio a wearable actually encounters.

\subsection{Base classifier}

The base model is a MobileNetV3 feature extractor with a multilayer-perceptron classification head, performing binary cough / not-cough classification, provided by the ARoS Laboratory at North Carolina State University. The model was used in inference mode only. No weights were retrained, no architecture was modified, and no fine-tuning was performed. Its operating point is a decision threshold of 0.7 on the cough probability. It is treated throughout as a fixed, frozen artifact, the experimental control against which all uncertainty methods are measured.

Audio is not consumed raw. Recordings are resampled to 32\,kHz, amplitude-normalized, segmented into fixed-length windows, and converted to mel-spectrogram representations, which the convolutional feature extractor consumes as images. Coughs appear in this representation as short, broadband, burst-like events; speech and environmental noise appear as more temporally continuous structures. When these patterns overlap in real-world audio, the model is more likely to make the errors studied in this paper.

\subsection{Uncertainty signals}
\label{sec:signals}

\paragraph{Monte Carlo Dropout.}
Dropout layers are enabled during inference, and $T$ stochastic forward passes are performed for each input. With $p_t$ representing the predicted cough probability on pass $t$, Eq.~\eqref{eq:mc} gives the predictive mean and variance:
\begin{equation}
  \bar{p} = \frac{1}{T}\sum_{t=1}^{T} p_t, \qquad
  \mathrm{Var}(p) = \frac{1}{T}\sum_{t=1}^{T} \left(p_t - \bar{p}\right)^2 .
  \label{eq:mc}
\end{equation}
Here, $p_t$ is the predicted cough probability from dropout pass $t$, $\bar{p}$ is the mean probability across the $T$ passes, and $\mathrm{Var}(p)$ is the predictive variance. Using binary entropy $H(p) = -p\ln p - (1-p)\ln(1-p)$, predictive entropy is calculated from $\bar{p}$ and measures total uncertainty. Aleatoric uncertainty is the mean of $H(p_t)$ across the individual predictions, also called expected entropy. Epistemic uncertainty is predictive entropy minus aleatoric uncertainty, corresponding to the mutual-information estimate. All three entropy-based measures are calculated from the same $T = 20$ dropout passes. Offline experiments use 20 passes; the wearable prototypes use 5 to 10 passes to reduce computation on edge hardware.

\paragraph{Temperature Scaling.}
A single scalar temperature $\tau > 0$ rescales the logits before the softmax:
\begin{equation}
  q_i = \frac{\exp(z_i/\tau)}{\sum_{j=1}^{K}\exp(z_j/\tau)} ,
  \label{eq:temp}
\end{equation}
where $z_i$ is the logit for class $i$, $\tau$ is the calibrated temperature, and $K$ is the number of classes; $q_i$ is the calibrated probability for class $i$. A value of $\tau$ above 1 softens the distribution (reducing confidence), while a value between 0 and 1 sharpens it. The value of $\tau$ is fit on a held-out validation split by minimizing negative log-likelihood. Because temperature scaling shifts the probability mass without changing the argmax, applying it under a fixed decision threshold above 0.5 silently alters the operating point; we therefore also re-optimize the decision threshold on validation data to maximize F1.

\paragraph{Mahalanobis distance.}
For input $x$, $f(x)$ is the penultimate-layer embedding, $\mu$ is the mean embedding of the validation true positives, and $\Sigma$ is their covariance matrix. The Mahalanobis distance is
\begin{equation}
  D_M(x) = \sqrt{\left(f(x) - \mu\right)^{\top} \Sigma^{-1} \left(f(x) - \mu\right)} .
  \label{eq:maha}
\end{equation}
A large distance indicates that the input lies far from the validation true-positive distribution in representation space. This suggests that it is unlike the correctly detected coughs used as the reference, regardless of the confidence assigned by the classifier.

\subsection{Experimental design}

The independent variable is the combination of uncertainty methods applied to the frozen classifier. The methods are added one at a time to measure the effect of each one: baseline (standard inference, no uncertainty quantification, temperature 1, threshold 0.7); plus Monte Carlo Dropout; plus Temperature Scaling with threshold recalibration; and plus the Mahalanobis filter (the full SECURE-AI framework). Dependent variables are precision, recall, F1, accuracy, confidence gap (accuracy minus mean confidence; zero indicates no average confidence bias but does not establish calibration across all confidence levels), false-positive rejection rate, and false-positive rate by error category. Controlled throughout: identical datasets and splits, preprocessing, frozen model weights, evaluation code, and hardware. Tuned parameters, optimized on validation data, are the temperature $T_{\mathrm{cal}} = 8.57$ (negative-log-likelihood-minimizing on the Coswara validation split) and the decision threshold 0.54 (F1-maximizing after calibration, recalibrated from 0.7). The Mahalanobis reference (mean and covariance) is fit on the validation true positives. Rather than fix a single rejection threshold, we characterize the full rejection/retention tradeoff on the test set.

The Coswara dataset was partitioned 50/50 into validation and test sets. Each set contained approximately the same proportion of cough and non-cough recordings, yielding a held-out test set of $n = 12{,}161$. The temperature, decision threshold, and Mahalanobis reference mean and covariance were estimated using only the validation set; the classifier weights remained unchanged. No single Mahalanobis rejection threshold was selected. Instead, the operating points reported in the Results were chosen from the labeled test-set distributions to illustrate the tradeoff between retaining true coughs and rejecting false alarms. Therefore, these points are descriptive examples and should not be treated as deployment-ready thresholds.

\subsection{Error taxonomy construction}

A subset of the misclassified test samples was manually reviewed by listening to the audio two to three times and assigning each to a category based on its acoustic content. Given the size of the Coswara dataset, reviewing the full misclassification set was not practical. The subset was intended to identify the dominant error types rather than to produce exact population counts. False positives and false negatives were categorized separately because they had different types of errors. This manual review was then used to determine which uncertainty signal detected each type of error.

\subsection{Statistical analysis}

Statistical significance of the separation between the Mahalanobis distance distributions for true positives and false positives was assessed using Welch's two-sample $t$-test, which does not assume equal variances across groups. It is an appropriate choice here, since the two distributions differ in spread as well as in location. Effect size is reported as Cohen's $d$. To measure how well Mahalanobis distance separates true positives from false positives, we calculate the area under the ROC curve (AUROC). In this analysis, false positives are treated as the cases the system should flag. Because AUROC does not depend on choosing a specific rejection threshold, it is used as the main measure of separation.

We also evaluated whether Mahalanobis distance could identify missed coughs among predictions classified as non-coughs. This analysis included 8,993 test samples: 805 false negatives and 8,188 true negatives. Separation was measured using the area under the ROC curve, with the false negative treated as the class to be flagged. Because the reference distribution is built from confirmed coughs, a missed cough was expected to lie closer to it than a genuine non-cough; the observed direction confirmed this.

\section{Results and Discussion}

\subsection{Baseline: high accuracy conceals low precision}

\begin{table}[t]
  \caption{Baseline performance, standard inference (threshold = 0.7, no uncertainty quantification).}
  \label{tab:baseline}
  \centering
  \begin{tabular}{lcccc}
    \toprule
    Dataset & Accuracy & Precision & Recall & F1 \\
    \midrule
    ESC-50  & 97.2\% & 40.0\% & 80.0\% & 53.3\% \\
    Coswara & 78.2\% & 50.6\% & 82.8\% & 62.8\% \\
    \bottomrule
  \end{tabular}
\end{table}

The ESC-50 result illustrates the limitation of accuracy on imbalanced data (Table~\ref{tab:baseline}). Accuracy of 97.2 percent appears strong, but coughs are only 2 percent of the dataset, so a classifier that never predicted cough would score 98 percent. Precision of 40 percent means three of every five detections are false. On class-imbalanced detection tasks, high accuracy can mainly reflect performance on the majority class while hiding poor detection of the minority class.

On Coswara, precision remains modest at 50.6 percent: about half of all cough detections are wrong. Analysis of the predicted-cough set shows that unfamiliar sounds alone account for over 40 percent of all cough predictions, exceeding the number of true coughs. False positives are therefore driven principally by unfamiliar audio rather than by confusion among known respiratory classes. This is the first indication that a method sensitive to input novelty, rather than to class confusion, is what is required.

\subsection{Confidence alone does not separate correct from incorrect}

Mean confidence differs across outcome groups. Correctly classified coughs and speech score high, unfamiliar samples somewhat lower. But the distributions overlap substantially, and confidence on unfamiliar sounds remains high in absolute terms. Raw softmax confidence is therefore not a usable rejection signal on this data: it does not reliably separate correct predictions from incorrect ones.

\subsection{Monte Carlo Dropout: uncertainty anti-correlates with error}

Monte Carlo Dropout ($T = 20$ passes) produces a mean prediction that closely tracks the standard baseline, so classification performance is essentially unchanged from Table~\ref{tab:baseline}. The uncertainty estimates it produces, however, are not merely uninformative but inverted: predictive entropy is lower for incorrect predictions than for correct ones (0.451 versus 0.571; incorrect predictions are about 21 percent lower). The consequence is visible in selective prediction. If an uncertainty estimate is sound, discarding the most uncertain predictions should raise accuracy on those retained. It does the opposite (Table~\ref{tab:selective}).

\begin{table}[t]
  \caption{Selective prediction under Monte Carlo Dropout uncertainty (Coswara test set, rejection by predictive entropy). Rejecting the highest-uncertainty samples degrades accuracy, indicating the estimate is anti-correlated with error. ESC-50 values are from a separate evaluation.}
  \label{tab:selective}
  \centering
  \begin{tabular}{lcc}
    \toprule
    Coverage & ESC-50 Accuracy & Coswara Accuracy \\
    \midrule
    100\% (baseline) & 97.2\% & 78.2\% \\
    50\%             & 95.3\% & 60.4\% \\
    \bottomrule
  \end{tabular}
\end{table}

Retaining only the 50 percent of Coswara predictions the model is most confident about drops accuracy by 17.8 points. The model's confidence is not just poorly calibrated; it is misleading. Many incorrect predictions carry low epistemic uncertainty and are therefore retained, while correct predictions are disproportionately discarded. This is the paper's first substantive finding: Monte Carlo Dropout does not detect this model's errors, specifically the false-positive ones. The errors that matter are cases where the model is stably, reproducibly, and wrongly certain, so the stochastic forward passes cannot see them, because there is nothing for the passes to disagree about.

\subsection{Temperature scaling: calibration improves, discrimination does not}
\label{sec:temp}

Applying temperature scaling under the fixed 0.7 threshold is actively harmful. Softening the probabilities pushes true coughs below the fixed threshold, so recall collapses. Only when the threshold is re-optimized on validation data alongside the temperature (Coswara validation set: temperature 8.57, threshold 0.54) does the calibrated operating point become coherent (Table~\ref{tab:temp}).

\begin{table}[t]
  \caption{Temperature scaling with threshold recalibration, Coswara held-out test set. Calibration improves markedly; discrimination does not.}
  \label{tab:temp}
  \centering
  \begin{tabular}{lccc}
    \toprule
    Metric & Original ($T{=}1$, th${=}0.7$) & Recalibrated ($T{=}8.57$, th${=}0.54$) & Change \\
    \midrule
    Accuracy               & 78.2\%  & 82.9\%  & $+4.7$ \\
    Precision              & 50.6\%  & 59.9\%  & $+9.3$ \\
    Recall                 & 82.8\%  & 70.2\%  & $-12.6$ \\
    F1                     & 62.8\%  & 64.7\%  & $+1.9$ \\
    Confidence gap         & $-0.1654$ & $-0.0262$ & -- \\
    High-confidence errors & 36.3\%  & 21.6\%  & $-14.7$ \\
    \bottomrule
  \end{tabular}
\end{table}

The calibration improvement is substantial: the confidence gap shrinks by roughly an order of magnitude and high-confidence errors fall by 14.7 points. However, F1 improves by only 1.9 points. Calibration rescales confidence but does not improve the model's ability to distinguish coughs from non-coughs. At this calibrated operating point (temperature 8.57, threshold 0.54), the held-out test set ($n = 12{,}161$) yields a confusion matrix of 1,899 true positives, 1,269 false positives, 805 false negatives, and 8,188 true negatives, a precision of 59.9 percent and recall of 70.2 percent. This matrix is the starting point for the embedding-space analysis that follows. Notably, the calibrated confidence does support useful selective prediction: at 50 percent coverage the recalibrated model gains over 11 accuracy points relative to baseline. Calibration is therefore necessary to make confidence rankable, but not sufficient, because it cannot see inputs that lie far from the reference distribution.

\subsection{Mahalanobis distance: the decisive signal}
\label{sec:maha}

\begin{figure}[t]
  \centering
  \includegraphics[width=0.72\linewidth]{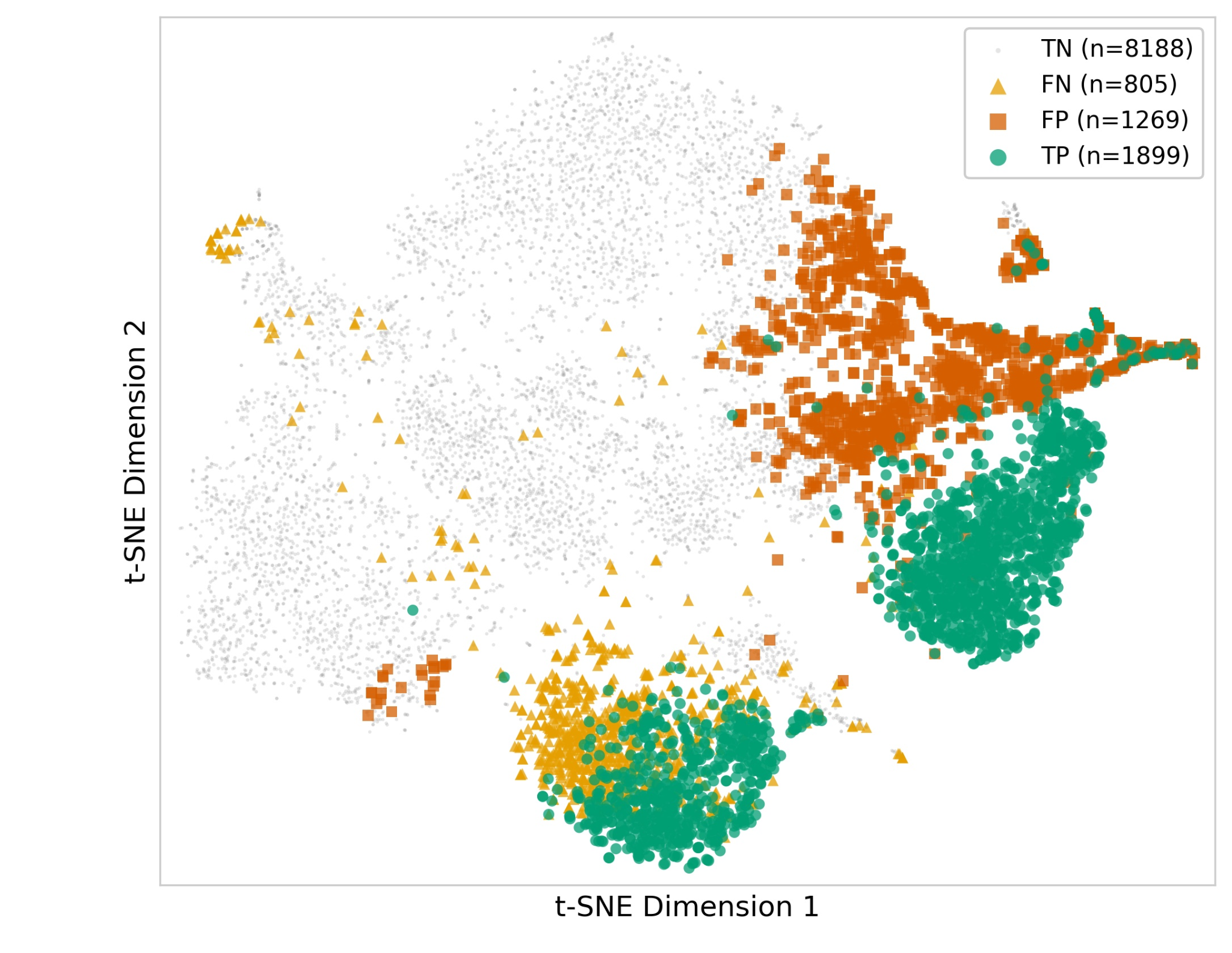}
  \caption{Penultimate-layer embedding space visualized with t-SNE, Coswara test set, colored and shaped by outcome: true negatives (grey dots, $n = 8188$), false negatives (orange triangles, $n = 805$), false positives (vermillion squares, $n = 1269$), and true positives (green circles, $n = 1899$).}
  \label{fig:tsne}
\end{figure}

Monte Carlo Dropout measures model disagreement and temperature scaling corrects confidence; both operate in probability space. Neither directly tests whether an input is unlike the reference cough distribution. Mahalanobis distance (Eq.~\eqref{eq:maha}) measures this directly in embedding space. Our approach adapts the class-conditional Gaussian method of Lee et al.~\cite{lee2018}: rather than fitting a Gaussian per class, we fit a single reference distribution on the validation true positives, the embeddings of correctly detected coughs, and score every test sample by its distance to that reference, without further fitting. On the Coswara test set, true positives and false positives occupy different regions of the embedding manifold (Figure~\ref{fig:tsne}). The distance is substantially lower for true positives (mean 44.7, standard deviation 13.8, $n = 1899$) than for false positives (mean 63.4, standard deviation 26.3, $n = 1269$). The separation is statistically significant under Welch's two-sample $t$-test (unequal variances), $t(1739.5) = -23.25$, $p < 0.001$ ($p = 2.3\times10^{-104}$), with an effect size of Cohen's $d = 0.94$ (95 percent confidence interval 0.87 to 1.02), a large effect by conventional thresholds~\cite{cohen1988}.

\begin{figure}[t]
  \centering
  \includegraphics[width=0.5\linewidth]{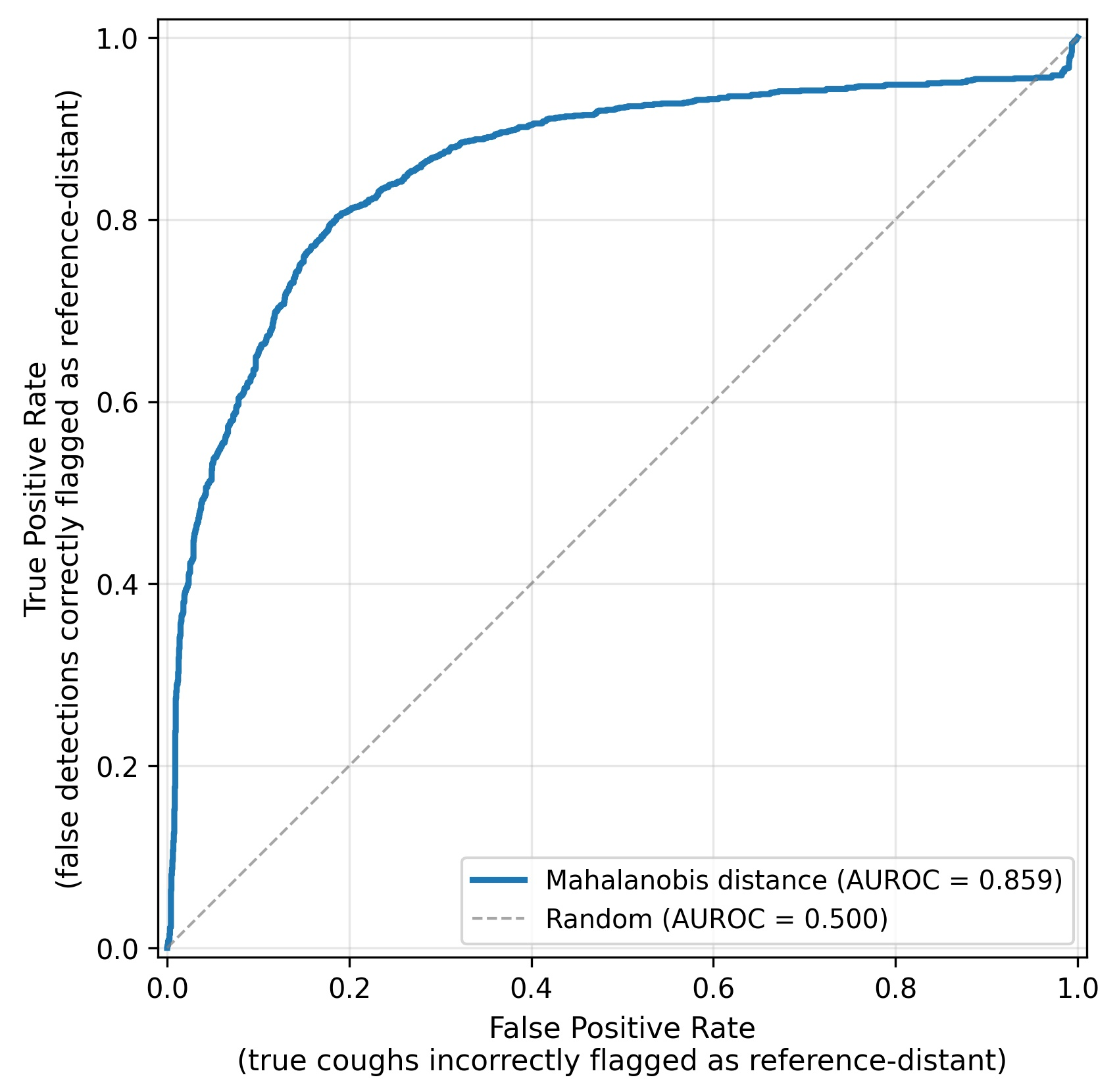}
  \caption{Receiver operating characteristic curve for Mahalanobis distance used to flag false-positive cough detections among predicted coughs (Coswara test set).}
  \label{fig:roc}
\end{figure}

Rather than commit to a single rejection threshold, we first quantify how well Mahalanobis distance separates true from false positives independently of any threshold, using the area under the ROC curve. Treating the false positive as the class to be flagged, Mahalanobis distance achieves an area under the curve of 0.86 (95 percent confidence interval 0.84 to 0.87, by bootstrap resampling) on the test set (Figure~\ref{fig:roc}). This measure does not depend on selecting a rejection threshold, so its result cannot be improved by selecting a favorable threshold. However, it still depends on the reference mean and covariance calculated from the validation data. In comparison, the two probability-based methods (Monte Carlo Dropout and Temperature scaling) did not clearly separate correct and incorrect predictions.

\begin{figure}[t]
  \centering
  \includegraphics[width=0.72\linewidth]{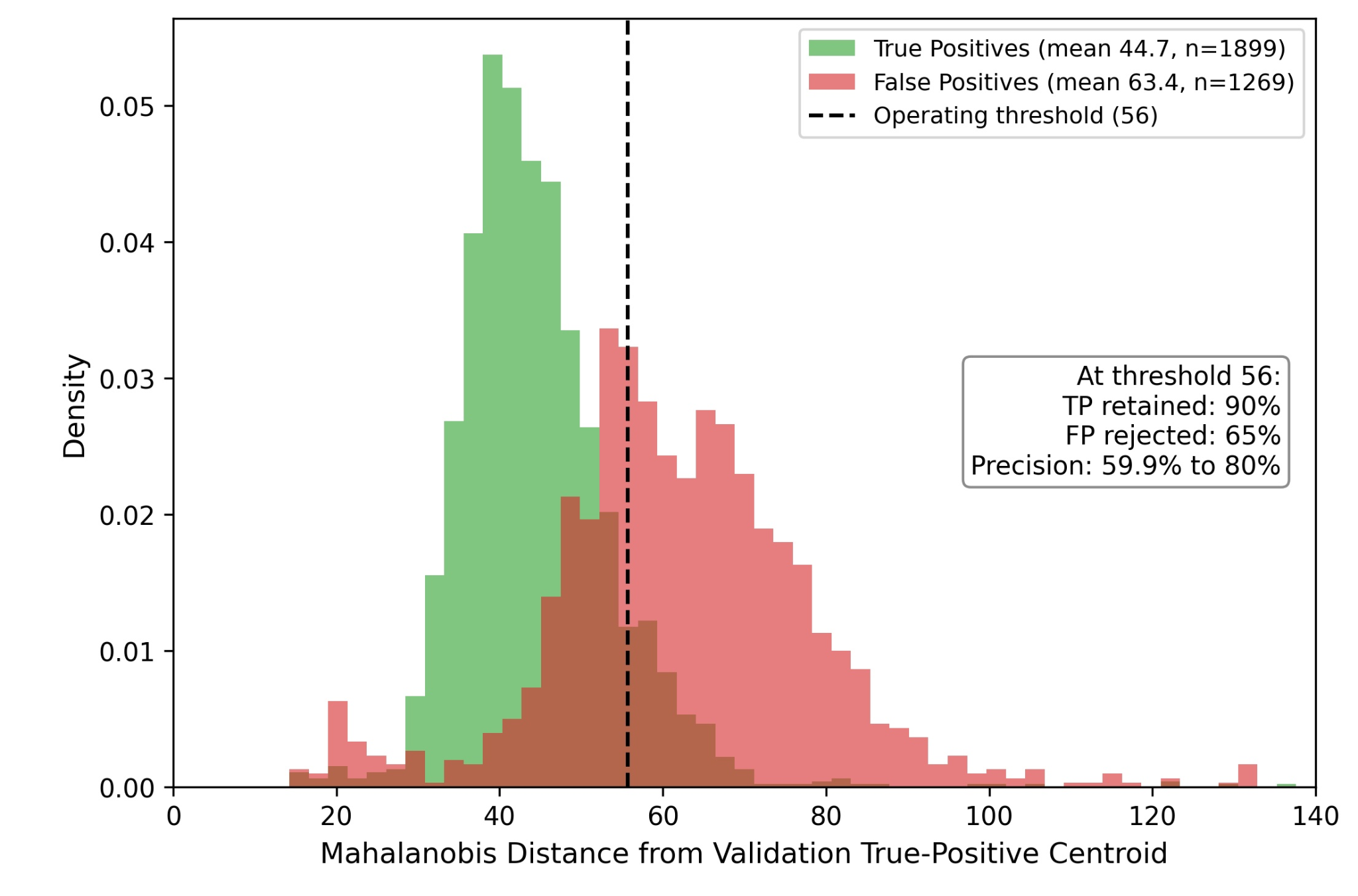}
  \caption{Distribution of Mahalanobis distances for true positives (mean 44.7) and false positives (mean 63.4) on the Coswara test set (Cohen's $d = 0.94$, Welch $t = -23.25$, $p < 0.001$; area under the curve 0.86).}
  \label{fig:hist}
\end{figure}

Every threshold involves a tradeoff between keeping true coughs and removing false alarms. Table~\ref{tab:operating} shows several examples from the test set. At the threshold that keeps 90\% of true coughs, about two-thirds of false alarms are removed, increasing precision from 59.9\% to about 80\%. Precision can be increased further, but more true coughs will be missed. Figure~\ref{fig:hist} shows the distance distributions for true and false positives and marks the threshold that keeps 90\% of true coughs.

\begin{table}[t]
  \caption{Representative operating points, read from the Coswara test-set Mahalanobis distributions. Each row is a point on the coverage-safety curve; no single row is privileged as the threshold. Precision is relative to the calibrated baseline of 59.9 percent.}
  \label{tab:operating}
  \centering
  \begin{tabular}{ccc}
    \toprule
    True coughs kept & False alarms removed & Precision \\
    \midrule
    98\% & 39\% & 71\% \\
    95\% & 53\% & 75\% \\
    90\% & 65\% & 80\% \\
    85\% & 76\% & 84\% \\
    80\% & 81\% & 86\% \\
    \bottomrule
  \end{tabular}
\end{table}

Mahalanobis distance also helped identify false negatives. Among the 8,993 predictions classified as non-coughs, it separated 805 false negatives from 8,188 true negatives with an area under the curve of 0.82. In this analysis, a lower distance meant that a recording was closer to the reference cough embeddings and was therefore more likely to be a missed cough. Thus, Mahalanobis distance provides information in both directions: unusually distant predicted coughs may be false positives, while unusually cough-like predicted non-coughs may be false negatives.

\subsection{The threshold does not transfer across splits}

A Mahalanobis threshold of 39.5 produced the highest F1 score on the validation set, with an F1 of 82.0\%. However, applying the same threshold to the test set reduced F1 to 34.3\%, a drop of nearly 48 percentage points (Table~\ref{tab:transfer}). This happened because the Mahalanobis reference was created from the validation set of true positives. As a result, validation true positives were closer to the reference center, with a mean distance of 30.8. Test true positives had a higher mean distance of 44.7. Therefore, the validation threshold was too low for the test set and incorrectly rejected many true coughs, causing recall to fall sharply.

\begin{table}[t]
  \caption{Validation-tuned threshold transfer. A threshold chosen to maximize F1 on validation performs well there and fails on test, because the validation reference distances are systematically compressed relative to test.}
  \label{tab:transfer}
  \centering
  \begin{tabular}{lcc}
    \toprule
     & Validation & Test \\
    \midrule
    Threshold (F1-optimal on validation)  & 39.5   & 39.5 \\
    F1 at that threshold                  & 82.0\% & 34.3\% \\
    Precision                             & 96.8\% & 88.6\% \\
    Recall                                & 71.1\% & 21.3\% \\
    Mean TP distance (reference offset)   & 30.8   & 44.7 \\
    \bottomrule
  \end{tabular}
\end{table}

This is the same phenomenon we observe for temperature (8.57 on Coswara versus 3.62 on ESC-50): uncertainty-calibration parameters, whether a scaling temperature or a distance-based rejection threshold, do not transfer across distributions, even between splits of the same dataset. This means that the separability of the signal (area under the curve 0.86, Figure~\ref{fig:roc}), not any single tuned threshold, is the honest summary of what embedding-space distance provides. A deployed system would need to calibrate its threshold on in-domain data from its own operating environment. This is a limitation, and a direction for future work.

\subsection{The combined system}

The reported rejection pipeline uses the temperature-scaled prediction at the calibrated threshold, followed by Mahalanobis rejection of predicted coughs. Monte Carlo Dropout is used to average predictions across stochastic passes, but its epistemic uncertainty is not used to reject predictions because it was nearly zero across all outcomes. Aleatoric uncertainty showed some separation between false negatives and true negatives, with lower values associated with false negatives, but Mahalanobis distance provided the stronger signal. Offline experiments use 20 dropout passes, while the wearable prototypes use 5 to 10 passes to reduce computation. Table~\ref{tab:combined} reports the system at the representative operating point that retains 90\% of true coughs. Relative to the calibrated baseline, precision rises from 59.9 percent to 79.5 percent and F1 from 64.7 percent to 70.4 percent, at the cost of recall falling from 70.2 percent to 63.2 percent.

\begin{table}[t]
  \caption{Combined system (temperature scaling plus Mahalanobis rejection) on the Coswara test set, at the 90-percent-true-cough-retention operating point. Baseline is the calibrated model before Mahalanobis rejection.}
  \label{tab:combined}
  \centering
  \begin{tabular}{lcc}
    \toprule
    Metric & Calibrated baseline & Combined system \\
    \midrule
    True positives  & 1,899 & 1,709 \\
    False positives & 1,269 & 440 \\
    False negatives & 805   & 995 \\
    True negatives  & 8,188 & 9,017 \\
    \midrule
    Precision & 59.9\% & 79.5\% \\
    Recall    & 70.2\% & 63.2\% \\
    F1        & 64.7\% & 70.4\% \\
    \bottomrule
  \end{tabular}
\end{table}

\subsection{Error taxonomy: failures are systematic}

Manual review of a subset of the misclassified test samples yields the distributions in Table~\ref{tab:taxonomy}. Over 61 percent of false positives are speech. Short, sharp vocal bursts of isolated letters or counting occupy the same acoustic neighborhood as coughs in the model's feature space. Failures are not random: they are concentrated, structured, and therefore addressable. False negatives tell a different story. The two dominant categories are nearly tied: regular, clear coughs (29.5 percent) and coughs obscured by background noise (27.3 percent). The clear-cough failures are especially notable, because these are acoustically unambiguous coughs that the model simply misses, a representation-level failure least amenable to the uncertainty methods studied here.

\begin{table}[t]
  \caption{Error taxonomy over a reviewed subset of misclassifications, Coswara test set ($n = 186$ false positives, $n = 176$ false negatives). Percentages are within-subset shares; the speech-related subtotal aggregates saying letters, counting, and speech interference. Percentages may not total 100\% because of rounding.}
  \label{tab:taxonomy}
  \centering
  \begin{tabular}{lrlr}
    \toprule
    False-positive category & Share & False-negative category & Share \\
    \midrule
    Saying letters         & 34.9\% & Regular cough sound & 29.5\% \\
    Counting               & 16.1\% & Background noise    & 27.3\% \\
    Speech interference    & 10.2\% & No coughs           & 12.5\% \\
    No sound               &  9.1\% & Weak amplitude      & 11.9\% \\
    Animal sounds          &  7.5\% & Not clear sounding  &  6.8\% \\
    Breathing sounds       &  5.9\% & Short duration      &  5.7\% \\
    Environmental impact   &  4.3\% & Atypical pattern    &  4.0\% \\
    Background noise       &  4.3\% & Multiple coughs     &  1.1\% \\
    Rhythmic sound         &  3.2\% & Other               &  1.1\% \\
    Respiratory non-cough  &  2.2\% &                     &        \\
    Other                  &  2.2\% &                     &        \\
    \midrule
    \textbf{Speech-related subtotal} & \textbf{61.3\%} & & \\
    \bottomrule
  \end{tabular}
\end{table}

\subsection{The diagnostic matrix: different errors, different signals}

The error taxonomy helps answer the main question of this paper: are all types of errors detected by the same uncertainty signal? The answer is no. Table~\ref{tab:diagnostic} compares each type of error with the uncertainty signal that detects it. Counting, isolated letters, and silence are sometimes incorrectly classified as coughs and have low epistemic and low aleatoric uncertainty. The model is confident in these predictions, so the probability-based methods do not flag them. However, these errors have high Mahalanobis distances and can be detected using the embedding-space method.

\begin{table}[t]
  \caption{Diagnostic matrix. Mahalanobis distance is expressed relative to correctly classified cases with the same prediction: false-positive rows are compared against true positives, false-negative rows against true negatives.}
  \label{tab:diagnostic}
  \centering
  \small
  \begin{tabularx}{\linewidth}{>{\raggedright\arraybackslash}X >{\raggedright\arraybackslash}p{1.6cm} >{\raggedright\arraybackslash}p{1.7cm} >{\raggedright\arraybackslash}p{2.6cm} >{\raggedright\arraybackslash}p{2.6cm}}
    \toprule
    Error class & Epistemic (MC Dropout) & Aleatoric & Mahalanobis (vs.\ same-prediction correct cases) & Detected by \\
    \midrule
    FN, background masking  & Near zero & Moderate--high & Not separately evaluated & Calibrated confidence \\
    FN, muffled / unclear   & Very low  & High           & Reduced                  & Mahalanobis + calibrated confidence \\
    FN, weak amplitude      & Very low  & High           & Reduced                  & Mahalanobis + calibrated confidence \\
    FN, regular clear cough & Low       & Low--moderate  & Reduced                  & Mahalanobis + calibrated confidence \\
    FP, counting speech     & Low       & Low            & Elevated                 & Mahalanobis only \\
    FP, saying letters      & Low       & Low--moderate  & Elevated                 & Mahalanobis only \\
    FP, silence             & Low       & Low            & Substantially elevated   & Mahalanobis only \\
    \bottomrule
  \end{tabularx}
\end{table}

False negatives show a different pattern. Coughs affected by background noise, muffling, or weak volume show moderate-to-high aleatoric uncertainty and lower calibrated confidence. Mahalanobis distance separated false negatives from true negatives with an area under the curve of 0.82, with lower distances indicating a missed cough (Section~\ref{sec:maha}). Overall, calibrated confidence detects ambiguous sounds, while Mahalanobis distance detects errors the model makes confidently. A system using either signal alone would miss the errors the other detects, and in a clinical setting the confident false positives are the more consequential omission.

\subsection{Edge deployment}

To test whether the framework is deployable rather than merely computable, two wearable prototypes were built (Table~\ref{tab:hardware}). Audio is captured through a digital I$^2$S MEMS microphone (INMP441-compatible) on the GPIO interface, buffered into fixed-length windows, and passed through the same preprocessing pipeline used offline. To conserve power, a two-tier strategy runs a lightweight acoustic screen continuously and triggers full model inference plus the uncertainty cascade only on candidate events. A detected cough that passes the Mahalanobis check is accepted, while a candidate event with elevated Mahalanobis distance is rejected. The pipeline runs on a 30-dollar device within a 512-megabyte memory budget, showing that the rejection mechanism is lightweight enough for wearable hardware. We report this as a feasibility demonstration; detailed latency, throughput, memory, power, and battery measurements were not systematically collected and are left for future work.

\begin{table}[t]
  \caption{Wearable prototypes. Both run the frozen classifier and the full detection pipeline on-device. Current-draw and battery-life values are approximate estimates, not timed measurements.}
  \label{tab:hardware}
  \centering
  \begin{tabular}{lll}
    \toprule
     & Prototype 1 & Prototype 2 \\
    \midrule
    Compute      & Raspberry Pi 4 (1.5\,GHz)   & Raspberry Pi Zero 2 W (1.0\,GHz) \\
    RAM          & 2--4\,GB                    & 512\,MB \\
    Sensors      & Audio only                  & Audio + 9-DOF IMU \\
    Power        & USB power bank (5\,V)       & 3.7\,V 3000\,mAh LiPo + boost \\
    Current draw & $\sim$700--900\,mA          & $\sim$250--400\,mA \\
    Battery life & $\sim$4--6\,h               & $\sim$6--10\,h \\
    Form factor  & Bench prototype             & Compact wearable \\
    Cost         & \$60--\$90                  & \$30--\$45 \\
    \bottomrule
  \end{tabular}
\end{table}

\subsection{Discussion}

Calibration is necessary but not sufficient. Temperature scaling improves calibration as expected: the confidence gap falls from $-0.165$ to $-0.026$ and high-confidence errors drop by 14.7 points. But F1 improves by only 1.9 points. A calibrated model produces confidence scores that can be trusted as a ranking, but calibration does not enable the model to recognize inputs unlike its training data. Because a lot of errors on this data originate in unfamiliar audio, this distinction is central to the task.

Confident errors are the failure mode that matters. The errors that survive both Monte Carlo Dropout and temperature scaling are the ones a clinical system can least tolerate: high confidence, low epistemic uncertainty, and incorrect. Silence classified as a cough with high softmax confidence would corrupt a patient's symptom record without any signal that the prediction is unreliable. The finding that these errors are separable in embedding space even when they are inseparable in probability space is the most transferable result of this work. No single method dominates, because the errors are heterogeneous. Ambiguous inputs produce high aleatoric uncertainty and are caught by calibrated confidence; novel inputs produce low uncertainty but high embedding distance and are caught only by the Mahalanobis test. Each signal detects a different class of failure, so a combination is required for coverage across both.

On limitations of this work, the base classifier was provided by the lab and used as-is, so this study analyzes its errors rather than fixing them; in particular, the 29.5 percent of false negatives that are clear coughs come from a weakness in the model's features that the framework can detect but not correct. Temperature scaling depends heavily on the dataset (8.57 on Coswara versus 3.62 on ESC-50), so calibration settings do not carry over between different audio conditions. The error taxonomy was labeled by a single person, so no inter-rater agreement was measured. The evaluation uses public datasets rather than clinically labeled recordings from real patients. Finally, no human-subjects data was collected, so this work shows what is needed for clinical use rather than demonstrating clinical use itself. Future work includes evaluation on clinically labeled respiratory audio; deep ensembles as an additional epistemic signal; migration to lower-power edge silicon; and, most importantly, using the finding that most false positives are speech to retrain the model on more speech examples, fixing the cause of these errors rather than filtering them post-hoc.

\section{Conclusion}

High benchmark accuracy does not imply safe real-world behavior, especially in class-imbalanced acoustic detection. The frozen cough classifier achieved 97.2\% accuracy on ESC-50 and 78.2\% on Coswara, but these accuracy values were strongly affected by class imbalance. Its cough-detection precision was modest on both datasets: 40.0\% on ESC-50 and 50.6\% on Coswara. Its errors were systematic rather than random, and confident false positives were not detected by the two widely used probability-space methods evaluated here. Monte Carlo Dropout helped describe the model's uncertainty patterns, but its epistemic signal did not reliably identify errors. Temperature scaling improved calibration without improving discrimination.

Mahalanobis distance provided the strongest signal. It separated true positives from false positives (Cohen's $d = 0.94$, Welch $t = -23.25$, $p < 0.001$) with an area under the curve of 0.86. Among predictions classified as non-coughs, lower Mahalanobis distance also separated false negatives from true negatives, with an area under the curve of 0.82. Thus, embedding distance helped identify missed coughs as well as false alarms. At a representative test-set operating point that kept 90\% of true coughs, Mahalanobis rejection removed roughly two-thirds of false positives and raised precision from 59.9\% to about 80\%. However, the validation-selected threshold did not transfer successfully to the test set: F1 fell from 82\% on validation to 34\% on test. Therefore, threshold-independent separability, rather than any single tuned threshold, is the most reliable measure of the information provided by embedding distance.

Overall, confidence calibration was helpful but was not sufficient on its own. Mahalanobis distance provided information in two directions: a predicted cough far from the reference cough distribution could be a false positive, while a predicted non-cough closer to the cough distribution could be a false negative. In a real deployment, the operating thresholds would need to be recalibrated using data from the environment in which the system will be used.

\section*{Ethics Statement}

This study used only publicly available, de-identified audio datasets (Coswara and ESC-50) and did not involve the collection of new data from human subjects or animals. No identifiable personal information was accessed, and no experiments involving human or animal participants were conducted by the author. As the work relied entirely on previously published open datasets, institutional review board approval and informed consent were not applicable.

\begin{ack}
The pre-trained MobileNetV3 cough detection model used as the base classifier in this work was provided by the Active Robotic Sensing (ARoS) Laboratory, Department of Electrical and Computer Engineering, North Carolina State University.\\ I thank Mrs.\ Mormando for her guidance, and my family for their support.
\end{ack}

\bibliographystyle{unsrtnat}
\bibliography{references}

\end{document}